%% file: arxiv.tex
\documentclass[10pt,twocolumn,letterpaper]{article}

\usepackage[pagenumbers]{cvpr} 

\input{preamble}

\usepackage[accsupp]{axessibility} 

\definecolor{cvprblue}{rgb}{0.21,0.49,0.74}
\usepackage[pagebackref,breaklinks,colorlinks,citecolor=cvprblue]{hyperref}

\def\paperID{6875} 
\def\confName{CVPR}
\def\confYear{2024}

\title{SPIDeRS: Structured Polarization for Invisible Depth and Reflectance Sensing}

\author{Tomoki Ichikawa \qquad
Shohei Nobuhara \qquad
Ko Nishino \\
Graduate School of Informatics, Kyoto University\\
{\tt\small https://vision.ist.i.kyoto-u.ac.jp/}
}

\begin{document}

\twocolumn[{
  \maketitle 
  \begin{center}
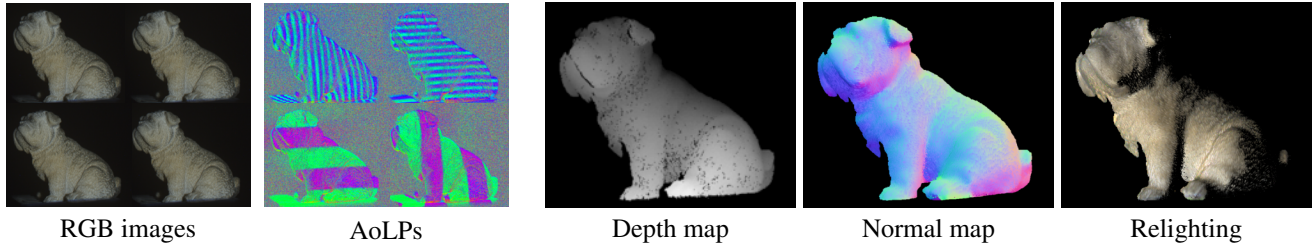

    \captionsetup{type=figure}

    \includetikzgraphics[opening]{fig_camera_ready/tikz_camera_ready}
    
    \captionof{figure}{
    We propose structured polarization, a novel invisible 3D sensing method using polarized light with per-pixel modulation of the angle of linear polarization. The method enables completely stealth measurement of 3D shape, surface normals, and reflectance.}
    \label{fig:opening figure}
\end{center}
}]

\begin{abstract}
Can we capture shape and reflectance in stealth? Such capability would be valuable for many application domains in vision, xR, robotics, and HCI. We introduce structured polarization for invisible depth and reflectance sensing (SPIDeRS), the first depth and reflectance sensing method using patterns of polarized light. The key idea is to modulate the angle of linear polarization (AoLP) of projected light at each pixel. The use of polarization makes it invisible and lets us recover not only depth but also directly surface normals and even reflectance. We implement SPIDeRS with a liquid crystal spatial light modulator (SLM) and a polarimetric camera. We derive a novel method for robustly extracting the projected structured polarization pattern from the polarimetric object appearance. We evaluate the effectiveness of SPIDeRS by applying it to a number of real-world objects. The results show that our method successfully reconstructs object shapes of various materials and is robust to diffuse reflection and ambient light. We also demonstrate relighting using recovered surface normals and reflectance. We believe SPIDeRS opens a new avenue of polarization use in visual sensing. 

\end{abstract}

\section{Introduction}
\label{sec:introduction}

Depth sensing has been a central topic of research in computer vision since its inception. Due to its many downstream applications, a variety of approaches have been explored. Structured light, in particular, is a practical method that achieves high accuracy, which has enjoyed commercial success in various forms. It exploits the duality of a projector with a camera and replaces one or more of a binocular or multiview stereo with a projector. By throwing known light patterns to the target, dense and accurate correspondences are established for triangulation. This is particularly useful for real-world objects which are often featureless and cannot be handled with camera stereopsis alone. Its ease of setup in addition to its dense reconstruction has made structured light a go-to approach in many applications, even for ground truth acquisition for 3D reconstruction research. 

Structured light, however, alters the appearance of the target as the pattern on the surface needs to be visible to the camera. This is undesirable for many applications outside the research lab or an industrial plant. For instance, such visible patterns would be distracting in xR systems. For this, many structured light implementations rely on infrared (IR) light and their extension to time-of-flight (ToF) sensing \cite{zhang2012microsoft,keselman2017intel} as IR light is invisible to the naked eye.
IR systems, however, cannot be used for recovering surface properties such as texture and reflectance as they are different in IR from those in the visual spectrum.

Can we make shape and texture capture invisible?
Is there a way to simultaneously recover the reflectance of the target such that we can even relight the object?
Such a system will make real-world object capture for downstream applications completely stealth, creating a large opportunity for effective communication (xR systems), advertisement, robotics, art, and nondestructive inspection, such as live scene editing and always-on change detection.

In this paper, we introduce Structured Polarization, a first-of-its-kind novel depth and reflectance sensing method using structured light of polarized light, which we refer to as SPIDeRS. The key idea is to project structured light patterns of varied angle of linear polarization (AoLP).
Polarization is invisible to the naked eye and a regular camera. By using structured polarization patterns and an RGB-polarimetric camera, we can realize invisible depth and texture capture. Polarization also gives us two advantages that none of the past visible or IR structured light methods have. The first is that the surface normals can be directly recovered providing dense fine details of the target surface. This is in sharp contrast to regular depth sensing where the normals can only be obtained from the triangulated depth, which is inevitably noisy due to the differentiation. The second is that we can also estimate the reflectance properties of the target surface from the polarimetric object appearance. This enables joint depth and reflectance sensing with a single set of invisible structured light patterns. 


We implement SPIDeRS by assembling a projector-camera system with an RGB-polarimetric camera and a polarization projector whose AoLP can be controlled in the throw at each pixel. For this, we leverage the fact that liquid crystal acts as a polarizer. We use a liquid crystal spatial light modulator (SLM) to modulate the polarization orientation of incident polarized light by altering the per-pixel voltage. This can be understood as an LCD projector without an analyzer that modulates the amplitude such that the AoLP instead of the intensity is modulated. To accurately decode the reflected AoLP structured pattern, we derive a novel method that can extract the pattern from the captured polarization image while accounting for ambient light and, most important, specular and diffuse reflection which encodes and is independent from the thrown AoLP, respectively. When there is no ambient light, the reflectance as the bidirectional reflectance distribution function (BRDF) and per-pixel surface normals can be recovered.

We validate SPIDeRS by evaluating its accuracy on a number of real-world objects. The results show that our method successfully reconstructs object shapes of various materials and is robust to diffuse reflection and ambient light. We also demonstrate successful reflectance and per-pixel surface normal recovery which can be used for relighting the object. We believe SPIDeRS opens a new avenue of research and practical use of polarization and will serve as a foundation for further studies.

\section{Related Work}
\label{sec:related work}

Various light patterns of intensity or color have been proposed for structured light sensing.
Salvi \etal~\cite{salvi2010state} review these structured light methods based on their coding strategies and requirements. Discrete spatial multiplexing methods project stripe or two-dimensional array patterns that encode codewords with neighboring pixels~\cite{boyer1987color,griffin1992generation}. Continuous patterns gradually change the intensity or wavelength of the projected light in space~\cite{carrihill1985experiments,tajima19903}.
Discrete time multiplexing methods separate the space by projecting binary or N-ary patterns~\cite{posdamer1982surface,caspi1998range}.
Phase shifting methods project sinusoidal stripe patterns with different phases and decode them by extracting them from captured images~\cite{srinivasan1985automated,gupta2012micro,moreno2015embedded}.
Frequency multiplexing methods project a sinusoidal or wavelet pattern with spatially varying phase shifts and decode them in the frequency domain~\cite{takeda1983fourier}.
Mirdehghan \etal~\cite{mirdehghan2018optimal} generate sequences of projection patterns that satisfy the specified conditions by minimizing the expected number of incorrect corresponding pairs.

Sundar \etal~\cite{sundar2022single} use a Single Photon Avalanche Diode (SPAD) array to capture binary structured light patterns.
Bajestani and Beltrame~\cite{bajestani2023event} use a structured light projector to capture color and depth with a monochrome event-based camera.
Xu \etal~\cite{xu2023unified} proposed a unified structured light to acquire both object shape and reflectance with an LCD mask and an LED array that enables angular sampling.
These methods use visible light which alters the appearance of the surface. This makes them visible and necessitates separate image acquisition for object appearance (\ie, texture).

Structured light with IR light avoids interference with the object appearance in the visual spectrum. It has been commercialized as Kinect v1~\cite{zhang2012microsoft, Smisek20113D} and has been applied to real-time applications and dense shape reconstruction from a single shot~\cite{ye2020accurate,schreiberhuber2022gigadepth,furukawa2022single}. These methods, however, also require separate visual spectrum image acquisition as IR light behavior is very different from that in the visual spectrum.

Huang \etal~\cite{huang2017target} proposed polarization-coded structured light for target enhancement.
It encodes binary structured light patterns as horizontally polarized light emitted from the green channel of an LCD projector and vertically polarized light emitted from the red and blue channels.
An ordinary camera with a polarizer rotated horizontally captures the patterns in intensity and decodes them.
This polarization-coded structured light is extended to capture in ambient light~\cite{huang2017polarimetric} and reconstruction in HDR~\cite{zhu2021rapid}.
These methods, however, project not only polarization but also color patterns which inevitably alter the surface appearance, \ie, they are visible and cannot capture texture.
Additionally, they can only project a binary pattern, which limits encoding strategies both in efficiency and accuracy.

In contrast, our method projects an imperceptible polarization pattern that retains the natural surface appearance and enables simultaneous texture acquisition with an RGB-polarimetric camera.
In addition to the 3D shape of the object, we can estimate its reflectance and surface normals without angular sampling by leveraging the captured surface appearance and polarimetric reflection.
Our method can use a continuous encoding pattern which provides sub-pixel accuracy by using a liquid crystal SLM.

\section{Polarization}
\label{sec:polarization}

Let us start by reviewing the properties of polarization~\cite{Hecht,ichikawa2021sky,fukao2021stereo}. 

\subsection{Basic Property of Polarization}
Light is a transverse electromagnetic wave oscillating perpendicularly to its propagation direction.
While linearly polarized light oscillates in a single orientation, circularly polarized light rotates its oscillating orientation over time.
Completely polarized light is generally represented as a superposition of coherent linearly and circularly polarized light, which is called elliptically polarized light.
Unpolarized light is light whose oscillating orientation changes completely randomly.
Light between unpolarized and polarized light is called partially polarized.

A polarizer enables observation of linearly polarized light by only transmitting light oscillating in a specific orientation.
If we observe partially polarized light through a polarizer whose filter angle $\phi_c$ is rotated, the observed intensity changes sinusoidally as a function of it
\begin{equation}
    I(\phi_c) = \overline{I}(1 + \rho_L\cos(2\phi_c - 2\phi_L)) \,,
    \label{eq:polarization filter}
\end{equation}
where $\overline{I}$ is the average intensity over the filter angle, $\rho_L$ represents the strength of linear polarization which is referred to as the Degree of Linear Polarization (DoLP), and $\phi_L$ represents the orientation of polarization, referred to as the Angle of Linear Polarization (AoLP).
With a linear polarizer, we cannot distinguish circularly polarized light from unpolarized light.

To recover the linear polarization state, we need to capture at least three images with different filter angles.
A polarimetric camera with four on-chip polarizers of different filter angles $\phi_c = 0$, $\frac{\pi}{4}$, $\frac{\pi}{2}$, and $\frac{3}{4}\pi$ for each pixel enables us to acquire the linear polarization state in a single shot.

\subsection{Mathematical Representation of Polarization}
There are two different concise mathematical representations for polarization: Jones calculus and Mueller calculus.
Jones calculus handles coherent completely polarized light and we use it to represent a liquid crystal SLM.
Mueller calculus handles incoherent partially polarized light and we use it to represent reflection on the surface.

In Jones calculus, the polarization state is expressed with a Jones vector $\VEC{E}$
\begin{equation}
    \VEC{E} = \begin{bmatrix} E_x \\ E_y \end{bmatrix} = \begin{bmatrix} E_{0x}\exp(j\phi_x) \\ E_{0y}\exp(j\phi_y) \end{bmatrix}\,,
\end{equation}
where $E_{0\{x,y\}}$ and $\phi_{\{x,y\}}$ are the amplitude and phase of the electric field for the \{x,y\}-axis, respectively, and $j$ is the imaginary unit.
Modulation of amplitude and phase is expressed with a Jones matrix $\VEC{J}$
\begin{equation}
    \VEC{E}_o = \VEC{J}\VEC{E}_i \,,
\end{equation}
where $\VEC{E}_i$ and $\VEC{E}_o$ are the Jones vectors of the input and output light.
A Jones matrix $\VEC{J}$ consists of complex numbers.

In Mueller calculus, the polarization state is expressed with a Stokes vector $\VEC{s}$
\begin{equation}
    \VEC{s} = \begin{bmatrix} s_0 \\ s_1 \\ s_2 \\ s_3 \end{bmatrix} =  \begin{bmatrix}
    I(0) + I(\frac{\pi}{2}) \\ I(0) - I(\frac{\pi}{2}) \\ I(\frac{\pi}{4}) - I(\frac{3}{4}\pi) \\ s_3
    \end{bmatrix} = \begin{bmatrix} 2\overline{I} \\ 2\overline{I}\rho_L\cos\phi_L \\ 2\overline{I}\rho_L\sin\phi_L \\ s_3 \end{bmatrix}\,,
    \label{eq:Stokes vector}
\end{equation}
where $s_0$ represents the intensity of light, $s_1$ and $s_2$ represent the linearly polarized component, and $s_3$ represents the circularly polarized component.
The DoLP and AoLP can be extracted from the Stokes vector as 
\begin{equation}
    \rho_L = \frac{\sqrt{s_1^2 + s_2^2}}{s_0} \,,\quad \phi_L = \frac{1}{2}\tan^{-1}\left( \frac{s_2}{s_1} \right) \,,
\end{equation}
respectively.
We can convert a Jones vector into a Stokes vector, except for the absolute phase, as
\begin{equation}
    \VEC{s} = \begin{bmatrix}
        E_{0x}^2 + E_{0y}^2 \\ E_{0x}^2 - E_{0y}^2 \\ 2E_{0x}E_{0y}\cos(\phi_x - \phi_y) \\ -2E_{0x}E_{0y}\sin(\phi_x - \phi_y)
    \end{bmatrix}\,.
\end{equation}
Polarization transformation by reflection is expressed by a Mueller matrix $\VEC{M}$ 
\begin{equation}
    \VEC{s}_r = \VEC{M}\VEC{s}_i \,,
    \label{eq:Mueller matrix}
\end{equation}
where $\VEC{s}_i$ and $\VEC{s}_r$ are Stokes vectors of incident and reflected light, respectively.

\section{Structured Polarization}
\label{sec:polarization structured light}

We project AoLP patterns with the polarization projector and decode the patterns reflected by the object surface to recover the object shape.

\subsection{Polarization Projector}
\label{sec:polarization projector}
We construct a polarization projector that can control the AoLP of projected light at each pixel.
We realize this with an SLM consisting of Twisted Nematic (TN) liquid crystal.
While such SLM can be often found in a regular intensity projector as an amplitude modulator combined with an analyzing polarization filter, our purpose fundamentally differs. We use it as an AoLP modulator without the analyzer in our polarization projector implementation. We could simply remove the front polarizer in an LCD projector to repurpose it as a polarization projector, but we found this process to be error-prone due to the tightly sealed lens system in commercial projectors. We instead build one from the ground-up.

Liquid crystal is a state of matter that simultaneously shows liquid-like fluidity and solid-like anisotropy.
TN liquid crystal has a helical structure of molecules put between alignment layers of perpendicularly oriented molecules.
Its helical anisotropy rotates the polarization orientation of the incident polarized light.
A voltage applied between the alignment layers causes the molecules to become parallel to the electric field and disturbs this helical structure, which weakens the rotation of the polarization orientation.
The SLM is a two-dimensional array of cells consisting of the TN liquid crystal and electrodes that can control the voltage and hence polarization orientation for each pixel.

Light modulation by TN liquid crystal can be expressed with a Jones matrix.
\begin{equation}
    \VEC{J} = \VEC{J}_{R}(-\alpha) \begin{bmatrix} \cos\gamma - j\frac{\beta}{\gamma}\sin\gamma & \frac{\alpha}{\gamma}\sin\gamma \\ -\frac{\alpha}{\gamma}\sin\gamma & \cos\gamma + j\frac{\beta}{\gamma}\sin\gamma \end{bmatrix} \,,
    \label{eq:Jones TNLC}
\end{equation}
where $\VEC{J}_{R}(-\alpha)$ is the rotation matrix of angle $\alpha$, $\alpha=\pi/2$ is the twist angle of the alignment layers, $\beta$ is the birefringence that depends on the applied voltage, and $\gamma=\sqrt{\alpha^2 + \beta^2}$. Since we consider only modulation of the polarization state, we omit the absolute phase of light~\cite{yamauchi1995optimization, davis1998polarization}.
The birefringence $\beta$ reaches $0$ as the applied voltage becomes large.
We can rewrite \cref{eq:Jones TNLC} as
\begin{equation}
    \VEC{J} = \frac{\alpha}{\gamma}\VEC{J}_{R}(-\alpha + \gamma) + \VEC{J}_{R}(-\alpha) A\begin{bmatrix}
        e^{-jB} & 0 \\ 0 & e^{jB}
    \end{bmatrix}\,,
    \label{eq:Jones TNLC re}
\end{equation}
where $A = \frac{1}{\gamma}\sqrt{(\gamma-\alpha)^2\cos^2\gamma + \beta^2\sin^2\gamma}$ and $B = \tan^{-1}\left\{ \beta\sin\gamma/\left((\gamma-\alpha)\cos\gamma\right) \right\}$.
The first term in \cref{eq:Jones TNLC re} represents the intended voltage-dependent rotation of linear polarization and the second term represents the elliptical polarization. 

\begin{figure}[t]
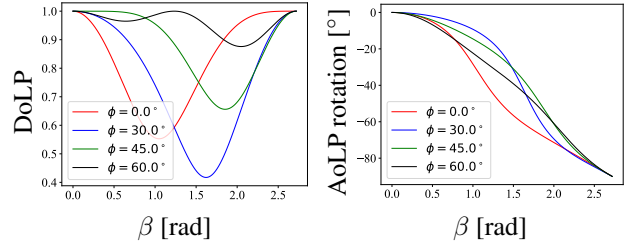

    \centering
    \includetikzgraphics[TNLC]{fig_camera_ready/tikz_camera_ready}
    \caption{DoLP and AoLP rotation through TN liquid crystal. The angle $\phi$ is the AoLP of the incident light. The TN liquid crystal enables us to control the AoLP arbitrarily in the range of $90^\circ$.}
    \label{fig:TNLC numerical analysis}
\end{figure}

When the applied voltage is small $\beta \gg \alpha$, the first term becomes $0$, $A=1$, and $B=\beta$.
In this case, we cannot control the rotation of polarization.
Otherwise the AoLP rotates as a function of the applied voltage.
\Cref{fig:TNLC numerical analysis} shows the DoLP and AoLP rotation of output light as a function of $\beta$ in the range of $\beta \leq \sqrt{3}\alpha$ for four AoLPs $\phi$ of incident light.
Although the DoLP decreases in some cases due to the circular polarization component, we can rotate the AoLP in the range of $90^\circ$ at will.

With an SLM and a polarized light source, we achieve per-pixel AoLP modulation in our projector throw.

\subsection{Polarimetric Reflection of AoLP Patterns}

To decode the projected AoLP patterns from its polarimetric reflection by the target object surface, let us consider how surface reflection modulates the polarization state of incident light.
Several polarimetric Bidirectional Reflectance Distribution Function (pBRDF) models that represent both specular and diffuse reflections have been proposed~\cite{Baek2018SimultaneousAO,kondo2020accurate,hwang2022sparse,ichikawa2023fresnel}.

Polarimetric specular reflection is derived from Fresnel reflection on mirror microfacets. Baek \etal~\cite{Baek2018SimultaneousAO} simplify the polarimetric specular reflection, assuming a co-axial projector-camera setup.
The simplified polarimetric specular reflection can be expressed with a Mueller matrix 
\begin{equation}
    M_s \approx c_s \mathrm{diag}(1,1,-1,-1) \,,
    \label{eq:polarimetric specular reflection}
\end{equation}
where $\mathrm{diag}(1,1,-1,-1)$ is a diagonal matrix and $c_s$ is a radiometric specular term that represents the shading and specular BRDF.
Since single scattering represents similar polarimetric behavior~\cite{hwang2022sparse}, we can incorporate it into \cref{eq:polarimetric specular reflection} by extending $c_s$ to the sum of the radiometric terms of the specular reflection and single scattering.

Specular reflection for a co-axial projector-camera retains the incident polarization state and is desirable for decoding polarization patterns.
For depth triangulation, however, the projector-camera system needs parallax, which causes change in the polarimetric state of specular reflection.
Fortunately, when the discrepancy of the viewing and lighting directions is less than $20^\circ$, AoLP rotation by specular reflection is theoretically less than $1.2^\circ$ and thus \cref{eq:polarimetric specular reflection} holds.
We strike this balance by locating the camera and the projector at a distance from the target surface and with enough distance from each other for triangulation.

Polarimetric diffuse reflection is derived from Fresnel transmission on a mirror macrofacet or microfacet. 
The transmitted light is depolarized through subsurface scattering and polarized again by re-transmission.
The Mueller matrix of diffuse reflection becomes
\begin{equation}
    M_d = c_d {\small \begin{bmatrix}
        1 & m_d^{12} & m_d^{13} & 0 \\ m_d^{21} & m_d^{22} & m_d^{23} & 0 \\ m_d^{31} & m_d^{32} & m_d^{33} & 0 \\ 0 & 0 & 0 & 0 \\ 
    \end{bmatrix}} \,,
    \label{eq:polarimetric diffuse reflection}
\end{equation}
where $c_d$ is the radiometric diffuse term that represents shading and diffuse BRDF and $m_d$ consists of the Fresnel transmittance of transmission into the surface and re-transmission into the air.
Baek \etal~\cite{Baek2018SimultaneousAO} assume that the elements of linear polarization modulation $m_d^{22}$, $m_d^{23}$, $m_d^{32}$, and $m_d^{33}$ are close to $0$ due to the small diffuse DoLP. We follow this assumption and represent polarimetric diffuse reflection as
\begin{equation}
    M_d \approx c_d {\small \begin{bmatrix}
        1 & m_d^{12} & m_d^{13} & 0 \\ m_d^{21} & 0 & 0 & 0 \\ m_d^{31} & 0 & 0 & 0 \\ 0 & 0 & 0 & 0 \\ 
    \end{bmatrix}} \,.
    \label{eq:polarimetric diffuse reflection2}
\end{equation}

The observed Stokes vector when the AoLP pattern is projected in ambient light becomes
\begin{equation}
    \VEC{s}_o = (M_s + M_d) \VEC{s}_i + \VEC{s}_{a} \,,
    \label{eq:observed stokes vector}
\end{equation}
where $\VEC{s}_i$ is the Stokes vector of projected light and $\VEC{s}_{a}$ is the Stokes vector of reflected ambient light.
From \cref{eq:polarimetric specular reflection,eq:polarimetric diffuse reflection,eq:observed stokes vector}, we can express the observed AoLP as
\begin{equation}
    \phi_o = \frac{1}{2}\tan^{-1}\frac{-c_s s_i^2 + c_d m_d^{31} s_i^0 + s_a^2}{c_s s_i^1 + c_d m_d^{21} s_i^0 + s_a^1}\,,
    \label{eq:observed AoLP}
\end{equation}
where $s_x^0$, $s_x^1$, and $s_x^2$ are the elements of the corresponding Stokes vector $\VEC{s}_x$.
The ratio of $s_i^2$ and $s_i^1$ represents the projected AoLP pattern.
Since polarimetric modulation of specular reflection is represented by a constant diagonal matrix, it retains the polarization of the projected pattern.
Diffuse reflection and ambient light, however, modulate the projected pattern and $\phi_o$ deviates from the incident AoLP $\phi_i = \frac{1}{2}\tan^{-1}\frac{s_i^2}{s_i^1}$.
In \cref{sec:encoding and decoding AoLP patterns}, we show that we can robustly extract the projected AoLP from the polarimetric observation \cref{eq:observed stokes vector}.

\begin{figure}[t]
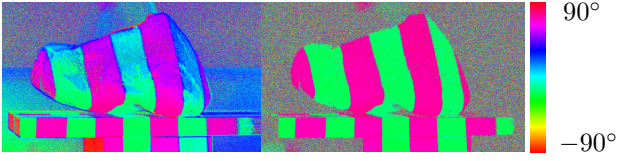

    \centering
    \includetikzgraphics[ambient]{fig_camera_ready/tikz_camera_ready}
    \caption{AoLP pattern extraction from polarimetric reflection on the object. Diffuse reflection and ambient light alter the AoLP throw in the reflected polarimetric object appearance and cause decoding errors (left). Our extraction method robustly extracts the true projected pattern (right).}
    \label{fig:remove ambient}
\end{figure}
\subsection{Encoding and Decoding AoLP Patterns}
\label{sec:encoding and decoding AoLP patterns}

We can encode the projector pixel with spatially varying AoLP values instead of intensity.
Since we can continuously change the AoLP at each pixel in the polarization projector throw in the range of $90^\circ$ as shown in \cref{sec:polarization projector}, we can realize any of the patterns used in regular structured light, such as time multiplexing and phase shifting.

We extract the AoLP of the projected light from polarimetric images represented with \cref{eq:observed stokes vector}.
In \cref{eq:observed AoLP}, the diffuse reflection affects $\phi_o$ only through $s_i^{0}$ while the specular reflection affects it through $s_i^{1}$ and $s_i^{2}$.
By subtracting the effect of $s_i^{0}$ from \cref{eq:observed stokes vector}, we can remove the polarized diffuse component from $\phi_o$ while retaining the polarized specular component.
To implement this diffuse removal, we use another polarimetric image of projected unpolarized light with the same intensity as $\VEC{s}_i$ whose observed Stokes vector becomes
\begin{equation}
    \VEC{s}_{\hat{o}} =
    \begingroup
    \setlength\arraycolsep{3pt}    
    \begin{bmatrix}
        c_s s_i^0 + c_d s_i^0 &
        c_d m_d^{21} s_i^0 &
        c_d m_d^{31} s_i^0 &
        0
    \end{bmatrix}\trans
    \endgroup
     + \VEC{s}_a \,.
    \label{eq:observed stokes vector under unpolarized light}
\end{equation}
By subtracting \cref{eq:observed stokes vector under unpolarized light} from \cref{eq:observed stokes vector}, we can obtain $\VEC{s}_{-} = \VEC{s}_o - \VEC{s}_{\hat{o}}$ as
\begin{equation}
    \VEC{s}_{-} =
    \begingroup
    \setlength\arraycolsep{2.5pt}    
    \begin{bmatrix}
        c_d(m_d^{12}s_i^1 + m_d^{13}s_i^2) &
        c_s s_i^1 &
        -c_s s_i^2 &
        -c_s s_i^3
    \end{bmatrix}\trans \,.
    \endgroup
    \label{eq:subtracted stokes vector}
\end{equation}
The AoLP of incident pattern $\phi_i$ is extracted as
\begin{equation}
    \phi_i = -\frac{1}{2}\tan^{-1}\frac{s_{-}^2}{s_{-}^1}\,.
    \label{eq:extract AoLP}
\end{equation}
Since the polarization projector cannot project unpolarized light, we synthesize the polarimetric image represented as \cref{eq:observed stokes vector under unpolarized light} by averaging two polarimetric images captured when the projected polarization orientations are perpendicular to each other with the same DoLP.

From \cref{fig:TNLC numerical analysis}, we can capture these polarimetric images when the rotation of AoLP is $0^\circ$ and $-90^\circ$, respectively.
\Cref{fig:remove ambient} shows the effectiveness of our AoLP extraction method. While the diffuse and ambient light modulates the observed AoLP, the extracted AoLP shows a clear pattern.
By decoding the extracted patterns, we can obtain dense correspondences between the camera and the projector and reconstruct the shape via triangulation.

\subsection{Calibration}

\begin{figure}[t]
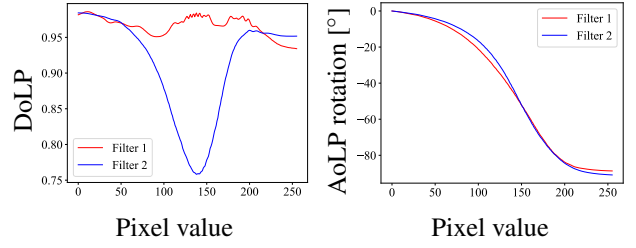

    \centering
    \includetikzgraphics[calibresult]{fig_camera_ready/tikz_camera_ready}
    \caption{Calibration results of DoLP and AoLP when the polarizing filter before the SLM is horizontal (Filter 1) and rotated by $45^\circ$ (Filter 2). We can continuously rotate the AoLP in the range of $[0^\circ, 90^\circ]$ by changing the pixel value. The DoLP is large enough to be detected for the horizontal filter.}
    \label{fig:calibration result}
\end{figure}

For accurate encoding, we polarimetrically calibrate our projector, \ie, establish the relation between the applied SLM pixel value and the polarization state of its actual throw. 
We put the polarimetric camera without a camera lens facing the polarization projector instead of a target object and directly acquire the polarization state of the projected light.
We input a uniform pattern of each pixel value into the SLM and obtain a DoLP and an AoLP from the average Stokes vector of the polarimetric camera.
\Cref{fig:calibration result} shows the calibration results of DoLP and AoLP of different incident polarized light.
The results show that we can continuously rotate AoLP in the range of $[0^\circ, 90^\circ]$ as shown in \cref{fig:TNLC numerical analysis}.
Since a larger DoLP is desirable for high-fidelity detection, we project horizontally polarized light onto the SLM.
We can use uniform patterns of $0$ and $255$ for the specular AoLP extraction.
Although the DoLP values for $0$ and $255$ are slightly different, \cref{fig:remove ambient} shows that this difference can be ignored.

We also calibrate the extrinsic and intrinsic parameters of the polarization projector for shape reconstruction, using known intrinsic camera parameters.
We first project the structured polarization light onto a ChArUco board of different poses and capture them.
By detecting the poses of the board, we obtain 3D points on the board at each camera pixel.
We also obtain corresponding points between the camera and projector by decoding the polarization pattern.
These correspondences provide 3D points on the board at some projector pixels, which enables us to calibrate the projector in the same way as camera calibration~\cite{zhang2000flexible}.

\begin{figure}[t]
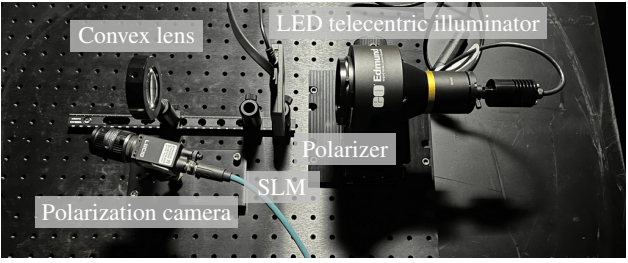

    \centering
    \includetikzgraphics[setup]{fig_camera_ready/tikz_camera_ready}
    \caption{Polarization projector-camera system implementing SPIDeRS. Our polarization projector consists of an LED telecentric illuminator with a polarizer, SLM, and convex lens.}
    \label{fig:experimental setup}
\end{figure}

\section{BRDF and Surface Normal Estimation}
\label{sec:BRDF and surface normal estimation}

If ambient light is not present, we can also estimate the surface normal and reflectance of a dielectric surface by exploiting their close relation in polarimetric reflection. When the incident polarization changes, the polarization of specular reflection changes but that of diffuse reflection remains the same. This enables the estimation of diffuse polarization and intensity of each reflection component from which we can recover the surface normals and BRDF, respectively.

We use FMBRDF~\cite{ichikawa2023fresnel}, which unifies polarimetric and radiometric reflectance of a dielectric surface, to represent the surface BRDF.
We assume that the diffuse (body reflection) albedo is spatially varying but the other FMBRDF parameters are uniform over the surface, \ie, the surface consists of a single material but has texture.
We use polarimetric images captured under uniform patterns for specular extraction to ensure reliable incident polarization states.

\vspace{-8pt}
\paragraph{Initial Estimates} We compute initial surface normals with Principal Component Analysis (PCA) of 3D points.
To obtain an initial estimate of the spatially varying diffuse albedo, we separate diffuse and specular reflections by using polarization.
From \cref{eq:Stokes vector,eq:subtracted stokes vector}, the specular reflection becomes
\begin{equation}
    I_s = c_s s_i^0 = \frac{1}{\rho_i}\sqrt{(s_{-}^{1})^2 + (s_{-}^{2})^2}\,,
    \label{eq:specular intensity}
\end{equation}
where $\rho_i$ is the DoLP of the projected light.
Temporarily assuming Lambertian diffuse reflection, we obtain initial diffuse albedo from diffuse reflection $I_d=s_o^{0} - I_s$.
We obtain initial estimates of the other uniform FMBRDF parameters by minimizing the error of intensity and DoLP
\begin{equation}
    \min_{\mu, k_s, \alpha, \beta, \kappa} \sum_i^N\sum_k^K  \left(L_{I,k,i} + \lambda_1 L_{\rho,k,i}\right) \,,
    \label{eq:initial estimation uniform paramter}
\end{equation}
where $L_{I,k,i}=|\overline{s}_{o,k,i}^0 - s_{o,k,i}^0|^2$ is the intensity error, $L_{\rho,k,i}=|\overline{\rho}_{o,k,i} - \rho_{o,k,i}|^2$ is the DoLP error, $\overline{s}_{o,k,i}^0$ and $s_{o,k,i}^0$ are the first elements of captured and rendered Stokes vectors at pixel $i$ of image $k$, respectively, $\overline{\rho}_{o,k,i}$ and $\rho_{o,k,i}$ are the captured and rendered DoLP, respectively, $N$ is the number of pixels, and $K$ is the number of images.
The optimized FMBRDF parameters are the refractive index $\mu$, specular albedo $k_s$, surface roughness $\alpha$, shape of microfacet distribution function $\beta$, and concentration parameter of the microfacet correlation function $\kappa$.
We set $\lambda_1 = 0.1$.

\begin{figure*}[t]
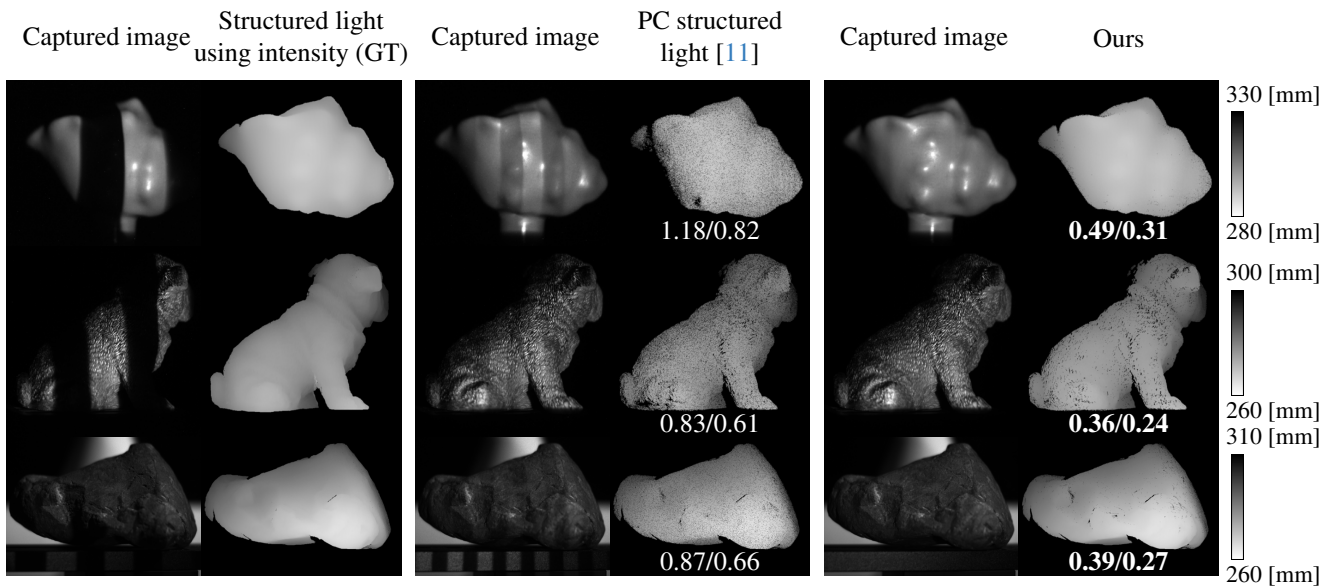

    \centering
    \includetikzgraphics[depthmap]{fig_camera_ready/tikz_camera_ready}
    \caption{Reconstructed depth maps of real objects of various materials and colors. The scene of the first and second rows do not have ambient light and the third row does. The numbers below each depth map are the mean and median of the depth errors in millimeters. The reconstruction results of our method are comparable with classic structured light, but its capture is completely invisible.}
    \label{fig:reconstructed depth map}
\end{figure*}

\begin{figure*}[t]
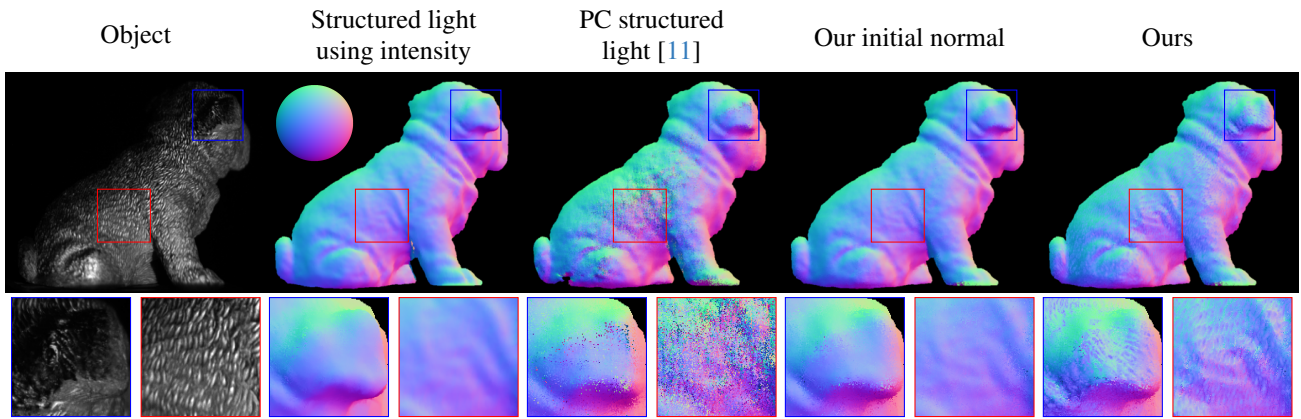

    \centering
    \includetikzgraphics[normalmap]{fig_camera_ready/tikz_camera_ready}
    \caption{Reconstructed surface normal maps of a real object. Our method fully exploits polarimetric reflection to directly estimate detailed surface normals instead of obtaining them as byproducts of depth values.}
    \label{fig:reconstructed normal map}
\end{figure*}

\begin{figure}[t]
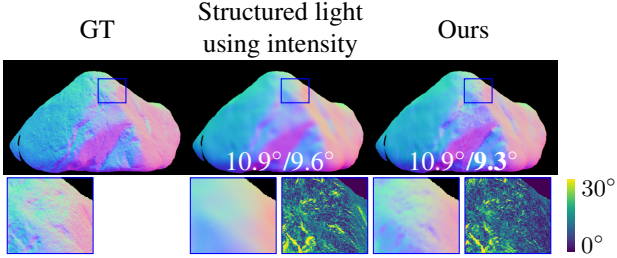

    \centering
    \includetikzgraphics[quantitative]{fig_camera_ready/tikz_camera_ready}
    \caption{Quantitative comparison of reconstructed normal maps of a real object. The numbers below each normal map are the mean and median of angular errors in degrees. Our method can accurately reconstruct detailed geometry.}
    \label{fig:normal map quantitative evaluation}
\end{figure}

\vspace{-8pt}
\paragraph{Joint Estimation of BRDF and Surface Normals} After the initial estimation, we refine the BRDF parameters and surface normals with joint optimization.
Since the AoLP depends on the surface normal, we add the error of the second and third elements of the Stokes vector to \cref{eq:initial estimation uniform paramter}
\begin{equation}
    \min_{\mu, k_s, \alpha, \beta, \kappa, \VEC{K}_b, \VEC{N}} \sum_i^N\sum_k^K \left( L_{I,k,i} + \lambda_2 L_{\rho,k,i} + \lambda_3 L_{s,k,i} \right) \,,
\end{equation}
where $L_{s,k,i}=|\overline{s_o^1} - s_o^1|^2 + |\overline{s_o^2} - s_o^2|^2$, $\VEC{K}_b$ represents the spatially varying diffuse albedo and $\VEC{N}$ represents surface normals.
We set $\lambda_2=0.01$ and $\lambda_3=1$.
For each pixel, we obtain six different equations of the Stoke vectors from two polarimetric images.
Since the unknown parameters are three pixel-wise and five global parameters, we have a sufficient number of equations to solve for them.

\section{Experimental Results}
\label{sec:experimental results}

We evaluate the accuracy of SPIDeRS with various real objects of different materials.
We also show that, in the absence of ambient light, we can estimate the BRDF and per-pixel surface normals of the object. 
We demonstrate object relighting with the estimated BRDF and surface normals.

As shown in \cref{fig:experimental setup}, we implement SPIDeRS with an LED telecentric illuminator with a polarizer, a spatial light modulator with a resolution of $1024{\times}768$ (HOLOEYE LC 2012), and an f=100mm convex lens.
We capture polarimetric images of the objects onto which our structured light is projected with a commercial monochrome polarimetric camera (Lucid TRI050S-PC) or a quad-Bayer RGB-polarimetric camera (Lucid TRI050S-QC).
Since polarization modulation of the SLM depends on wavelength, we use a green LED and the monochrome polarimetric camera for depth and normal reconstruction in \cref{sec:depth accuracy,sec:surface normal accuracy}.
For BRDF estimation in \cref{sec:relighting}, we use a white LED and the RGB-polarimetric camera, reconstruct the depths from the green channel, and estimate the surface normal and the BRDF for each channel.
Implementation of an RGB-polarization projector is left for future work. 

For the structured polarization patterns, we use phase-shifting unwrapped with Gray codes.
We post-process the decoding result by removing camera pixels whose corresponding projector pixels are discontinuous.
To alleviate failures of decoding on the edge (boundary) of Gray codes, we also project two sequences of shifted patterns and unify the decoding results of three sequences by averaging the positions of corresponding projector pixels.

\subsection{Depth Accuracy}
\label{sec:depth accuracy}

We first evaluate recovered depth map accuracy of SPIDeRS. 
We compare our method with classic structured light using intensity and polarization-coded (PC) structured light~\cite{huang2017target}.
For the classic structured light, we use the same encoding strategy as our method and project amplitude modulation patterns by inserting a polarizing filter between the SLM and the convex lens.
Since PC structured light only uses binary patterns, we emulate it with Gray code using our polarization projector.
As PC structured light uses an ordinary camera through a polarizing filter, we use one of the four filter angles of the polarimetric camera to mimic it.
Note that the original work uses an LCD projector and needs to project the patterns with color and thus fundamentally alters the object appearance.

\Cref{fig:reconstructed depth map} shows captured images and reconstructed depth maps of our method and other methods, with and without ambient light.
Holes in the reconstructed depth maps are caused by incorrect correspondences.
PC structured light suffers from decoding errors due to the low contrast of the patterns resulting from DoLP reduction and AoLP alteration by diffuse and ambient light and, by construction, cannot capture surface texture simultaneously due to the polarizing filter in front of the camera.
In contrast, our method successfully reconstructs object shapes of different materials and is robust to diffuse and ambient light while simultaneously capturing the surface appearance.
The results also show that the accuracy of our method is comparable with classic structured light while retaining the surface appearance and being totally invisible to the naked eye. Please see the supplementary material for results for a metallic surface.

\subsection{Surface Normal Accuracy}
\label{sec:surface normal accuracy}

\begin{figure}[t]
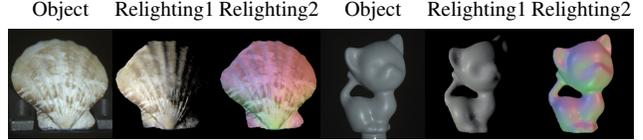

    \centering
    \includetikzgraphics[relighting]{fig_camera_ready/tikz_camera_ready}
    \caption{Relighting under a novel illumination with the estimated BRDF and surface normal.
    Illuminations are white directional lights from the left (Relighting1) and red, green, and blue ones from different directions (Relighting2).
    Our method can simultaneously estimate reflectance and surface normals along with a depth map by exploiting the captured texture and polarimetric reflection. The results can be used to relight the object.}
    \label{fig:relighting}
\end{figure}

\Cref{fig:reconstructed normal map,fig:normal map quantitative evaluation} show qualitative and quantitative comparisons of the estimated surface normal map and those computed from the reconstructed depth maps by PCA, respectively. We obtain the ground truth normal map by photometric stereo.
Our method successfully exploits polarimetric reflection to reconstruct detailed surface normals unlike past methods that can only produce them as differentiated byproducts of the measured depth maps.

\subsection{Relighting}
\label{sec:relighting}

\Cref{fig:relighting} shows relit target objects with estimated BRDFs under novel illuminations of different directions and colors.
Our method can estimate the BRDF accurately which results in qualitatively plausible relighting. This is possible as our method can simultaneously capture the texture and polarimetric reflection in the visible spectrum.
Past structured light methods cannot estimate the BRDF and surface normals as they only acquire radiometric images and the estimation is ill-posed.
Please see the supplementary material for more results.

\section{Conclusion}
\label{sec:conclusion}

In this paper, we introduced SPIDeRS, the first structured polarization depth and reflectance sensing method. SPIDeRS acquires object shape, reflectance, and surface normals from polarimetric images captured by modulating per-pixel AoLP in the projector throw. 
A liquid crystal SLM was used to achieve this structured polarization. 
We derived a novel method for robustly extracting the projected AoLP pattern from the observed polarimetric image formed by polarimetric reflection on the object surface. 
Our method enables object shape reconstruction while retaining the surface appearance intact so that surface texture and polarimetric reflection can simultaneously be captured from which the BRDF and surface normals can also be recovered. Most important, the whole process is completely invisible to the naked eye. 
We believe SPIDeRS provides a new means of depth and reflectance sensing and will serve as a foundation for further research in this exciting new area of structured polarization.

\vspace{-8pt}
\paragraph*{Acknowledgement}
This work was in part supported by
JSPS 
20H05951, 
21H04893, 
23H03420, 
and 23KJ1367, 
JST JPMJCR20G7 
and JPMJAP2305, 
and RIKEN GRP.

{
    \small
    \bibliographystyle{ieeenat_fullname}
    \bibliography{main}
}

\clearpage
\setcounter{page}{1}
\maketitlesupplementary

\appendix
\def\thesection{A.\arabic{section}}
\renewcommand\thefigure{A.\arabic{figure}}
\renewcommand\thetable{A.\arabic{table}}
\renewcommand\theequation{A.\arabic{equation}}
\setcounter{figure}{0}
\setcounter{table}{0}
\setcounter{equation}{0}

\section{Depth Accuracy of Metallic Surface}
\label{sec:depth accuracy of a metallic surface}

\begin{figure}[t]
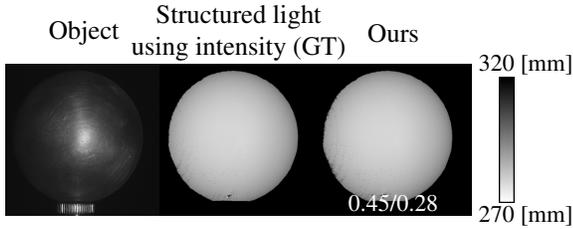

    \centering
    \includetikzgraphics[metallic]{fig_camera_ready/tikz_camera_ready}
    \caption{Reconstructed depth maps of a rough metallic surface. The numbers below each depth map are the mean and median of the depth errors in millimeters. Our method also works well for metallic surface.}
    \label{fig:metallic surface}
\end{figure}

\Cref{fig:metallic surface} shows the reconstructed depth maps for a rough metallic surface. While pure specular metallic surfaces would be challenging for any projector-camera system as the patterns will not be observed, for a rough metallic surface, our method successfully reconstructs the depth map. The proposed BRDF estimation method, however, is limited to dielectric materials as it uses the FMBRDF model~\cite{ichikawa2023fresnel}.

\section{Additional Experimental Results}
\begin{figure*}[t]
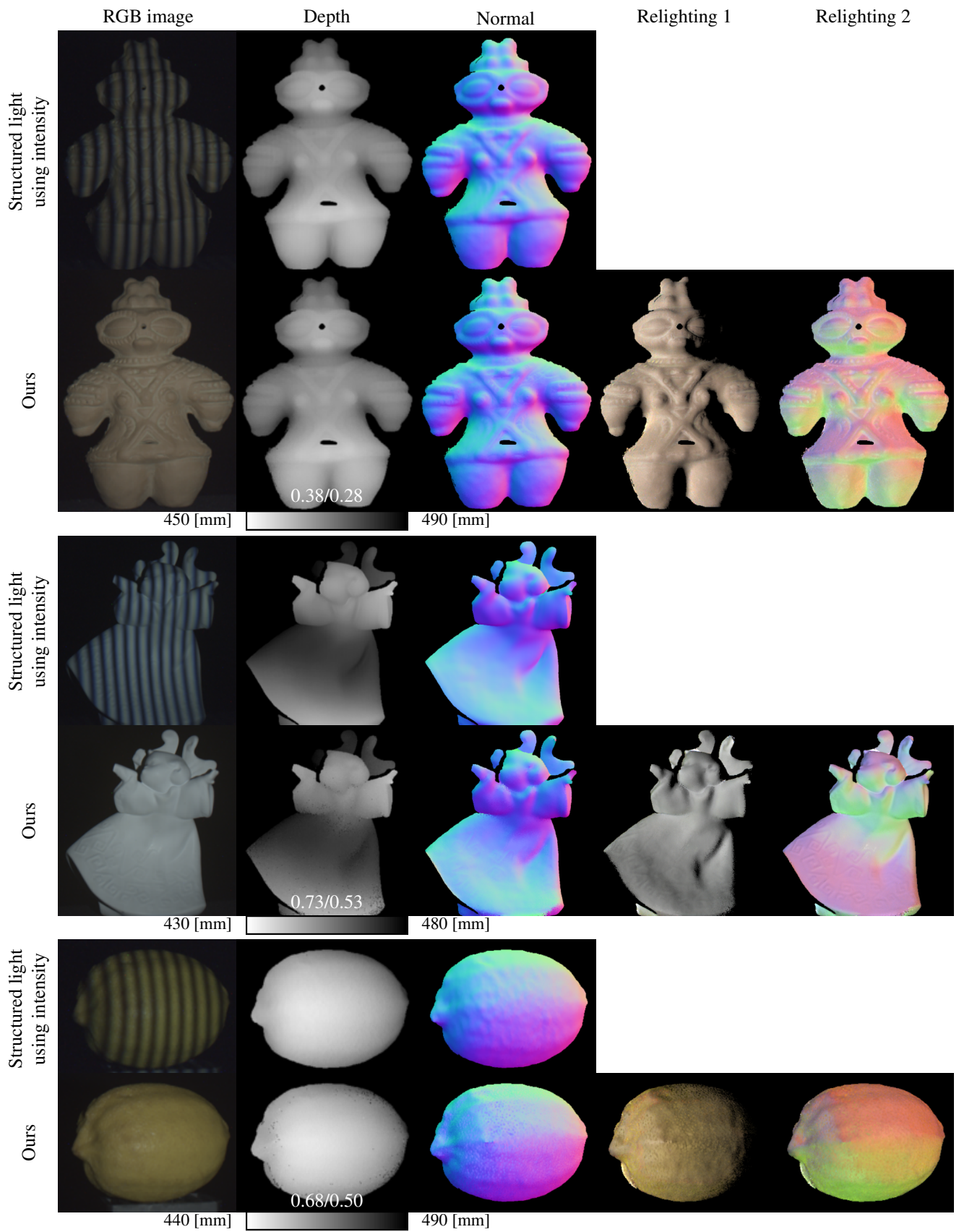

    \centering
    \includetikzgraphics[addresult]{fig_camera_ready/tikz_camera_ready}
    \caption{Additional results. The numbers below each depth map are the mean and median of the depth errors in millimeters.}
    \label{fig:additional results}
\end{figure*}

\Cref{fig:additional results} shows additional experimental results of depth and surface normal reconstruction and relighting.
The results show that the accuracy of the depth map reconstructed by our method is comparable with that of structured light using intensity while the capture process is invisible to the naked eye.
Our method also fully exploits polarimetric reflection to estimate the per-pixel surface normal and the BRDF, which we can use for qualitatively plausible relighting.

\section{Limitations}
Our analysis of polarimetric reflection is limited to ambient light and local illumination of the polarization projector. Diffuse inter-reflection is not affected by the projected polarization pattern and can be removed together with the diffuse reflection and ambient light. Specular inter-reflection depends on the pattern and can cause decoding errors. Such global illumination effects would also degrade the accuracy of normal and BRDF estimation. Rectifying the effects of global illumination with a spatial polarization pattern would be an important future work.

As mentioned in \cref{sec:depth accuracy of a metallic surface}, the normal and BRDF estimation is limited to dielectric materials due to the BRDF model. Polarimetric reflection on other materials, however, will also provide rich cues for a surface. By modeling analytical polarimetric and radiometric BRDF models on other materials, we will be able to extend our method to an even wider range of object surfaces.

\end{document}

%% file: preamble.tex
\usepackage[dvipsnames]{xcolor}

\usepackage{graphicx}
\usepackage{amsmath}
\usepackage{amssymb}
\usepackage{booktabs}
\usepackage{todonotes}
\usepackage{tabularx}
\usepackage{multirow}
\usepackage{multicol}
\usepackage{cite}
\usepackage{bm}
\usepackage{url}
\usepackage{balance}
\usepackage{enumitem}
\usepackage{color}
\usepackage{bbding}
\usepackage{nccmath}
\usepackage{adjustbox}
\usepackage{tikzinclude}

\def\VEC#1{{\boldsymbol{#1}}}

\def\trans{^\textsf{T}}